\documentclass[twoside]{article}

\usepackage[accepted]{aistats2025}

\usepackage{times} 
\usepackage{courier}  
\usepackage[hyphens]{url}  
\usepackage{graphicx} 

\usepackage[round]{natbib}
\usepackage{caption} 
\usepackage{algorithm}
\usepackage{algpseudocode}

\usepackage{microtype}
\usepackage{booktabs} 

\usepackage[utf8]{inputenc} 
\usepackage{url}            
\usepackage{booktabs}       
\usepackage{amsfonts}       
\usepackage{nicefrac}       
\usepackage{microtype}      
\usepackage{amsmath,bm}
\usepackage{lipsum}
\usepackage{color}
\usepackage{subcaption}
\usepackage{multirow}
\usepackage{verbatim} 
\usepackage{paralist}
\usepackage{centernot}
\usepackage{mathtools}

\usepackage{color}
\newlength\mylen

\usepackage{amsthm,amssymb}

\newtheorem{proposition}{Proposition}
\newtheorem{theorem}{Theorem}

\DeclareMathAlphabet\mathbfcal{OMS}{cmsy}{b}{n}

\newcommand{\blank}{{\color{white}.}} 

\newcommand{\V}[1]{\boldsymbol{\mathbf{#1}}}

\newcommand{\tuple}[1] {\langle #1 \rangle}
\newcommand{\eq}{\leftarrow}
\newcommand{\KL}[2]{D_{\text{KL}}({#1} \, || \, {#2})}
\newcommand{\x}{{\V{x}}}
\newcommand{\y}{{\V{y}}}
\newcommand{\R}{{\mathbb{R}}}
\newcommand{\X}{{\mathcal{X}}}
\newcommand{\PMC}{{P^{\text{MC}}}}

\definecolor{CmntColor}{rgb}{0.38, 0.30, 0.30}
\newcommand{\cmnt}[1]{{\color{CmntColor}$\triangleright$ \emph{\small{#1}}}}

\docsvlist{A, B, C, D, E, F, G, H, I, J, K, L, M, N, O, P, Q, R, S, T, U, V, W, X, Y, Z}

\newtheorem{corollary}{Corollary}

\newcommand*{\QED}{\null\nobreak\hfill\ensuremath{\square}}%

\begin{document}

\twocolumn[

\aistatstitle{Optimal Particle-based Approximation of Discrete Distributions (OPAD)}

\aistatsauthor{Hadi Mohasel Afshar \And Gilad Francis \And  Sally Cripps }

\aistatsaddress{Human Technology Institute, The University of Technology Sydney, New South Wales, Australia } ]

\begin{abstract}
Particle-based methods include a variety of techniques, such as Markov Chain Monte Carlo (MCMC) and Sequential Monte Carlo (SMC), for approximating a probabilistic target distribution with a set of weighted particles. 
\\
In this paper, we prove that for any set of particles, there is a unique weighting mechanism
that minimizes the Kullback-Leibler (KL) divergence of the (particle-based)
approximation from the target distribution, when that distribution is discrete -- 
any other weighting mechanism 
(e.g. MCMC weighting that is based on particles' repetitions in the Markov chain) is sub-optimal with respect to this divergence measure. 
Our proof does not require any restrictions either on the target distribution, or the process by which the particles are generated, other than the discreteness of the target. 
\\ 
We show that the optimal weights can be determined based on values that any existing particle-based method already computes;
As such, with minimal modifications and no extra computational costs, the performance of any particle-based method can be improved. 
Our empirical evaluations are carried out on 
important applications of discrete distributions including Bayesian Variable Selection and Bayesian Structure Learning.
The results illustrate that our proposed reweighting of the particles improves any particle-based approximation to the target distribution consistently and often substantially.  
\end{abstract}

\section{Introduction}
\label{sect.intro}

Discrete distributions arise in various scientific and engineering models \citep{chattamvelli2022discrete}. In these models, the number of states (i.e., configurations) often grows exponentially or even super-exponentially as the number of variables (dimensions) increases \citep{Kuipers2017}. This curse of dimensionality hinders the feasibility of exact inference. Consequently, high-dimensional discrete distributions are typically approximated using Markov Chain Monte Carlo (MCMC) methods \citep{pakman2013auxiliary,titsias2017hamming} or, in some cases, Sequential Monte Carlo (SMC) techniques \citep{schafer2013sequential}.

Both MCMC and SMC methods can be considered special cases of particle-based approaches. These algorithms approximate a target distribution using a finite set of weighted \emph{particles}, i.e., distinct states, where all weights are positive and sum to 1.

Our main contribution in this paper is to address the following general problem:\
\textit{In a particle-based approximation of a discrete target distribution, where the particles are arbitrarily chosen and fixed, what weight should be assigned to each particle to minimize the KL divergence of the resulting approximation from the target?}

We prove that this problem has a unique solution:\
\textit{The KL divergence is minimized if and only if the weight assigned to each particle is proportional to its target probability.}

We define an \emph{Optimal Particle-based Approximation of a Discrete Distribution} (OPAD) as any particle-based method in which the weights of the particles are proportional to their target probabilities. By simply reweighting the particles, any existing particle-based approximation of a discrete target distribution can be converted into an OPAD, which, as our theorem indicates, provides a better approximation of the target in terms of KL divergence.

This result also applies to MCMC methods. An MCMC chain is equivalent to a particle-based method where the particles are the distinct states of the chain, and the weight of each particle is proportional to its frequency of occurrence in the chain. The corresponding OPAD for this MCMC chain consists of the same particles, reweighted with weights proportional to their target probabilities.

Note that under mild assumptions, such as ergodicity, when the number of MCMC samples tends to infinity, the Markov chain converges. In discrete settings, this implies that the number of repetitions of each distinct state in the Markov chain approaches a value proportional to the target probability of that state. In other words, in the limit, the weight that MCMC assigns to each particle tends to its OPAD counterpart. This provides an intuitive insight into why OPAD outperforms MCMC: In MCMC, particle weights tend to their optimal values as the number of sampling iterations approaches infinity, whereas in OPAD, the optimal weight is assigned to each particle the first time it is visited.

Constructing an OPAD does not incur any additional computational overhead. This is because existing particle-based methods already compute the score (i.e., unnormalized target probability) of each particle. For example, in MCMC, these scores are necessary to calculate the Metropolis-Hastings acceptance probabilities of the proposals. Therefore, OPAD simply requires storing these computed values.

Finally, we note that MCMC methods 
also compute the scores of some states that are not added to the Markov chain (i.e., the rejected proposals). Since our main theorem imposes no restrictions on the choice of particles, we can include these rejected proposals as OPAD particles. This leads to a new variant of OPAD, which we call OPAD+, that can contain more particles than the original OPAD. We prove that the KL divergence of OPAD+ from the target is less than that of OPAD, just as the KL divergence of OPAD from the target is less than that of MCMC or any other corresponding particle-based method.

To quantify the significance of these theoretical improvements, we conduct experiments on the following well-known discrete target distributions: (a) the Ising model; (b) the posterior distribution of the selection indicator vector in a Bayesian Variable Selection model (using both synthetic and real-world observational data); (c) the posterior distributions of Directed Acyclic Graphs (DAGs) representing Bayesian Networks in the Bayesian Structure Learning setting.
Our results show that for all experimental models, the KL divergence of OPAD/OPAD+ from the target is significantly lower than that of the corresponding MCMC methods, by an order of magnitude ranging from one-tenth to one-thousandth.

\section{Optimal Particle-based Approximation of Discrete Distributions (OPAD)}
\label{sect.setting}

In this paper, we are interested in approximating a \emph{discrete target distribution}, $\pi^*$, 
that is defined on a countable but not necessarily finite \emph{support} set $\X$.
We assume $\pi^*$ can be evaluated up to a normalization constant. 
That is, we can evaluate a \emph{score function}, 
$\pi$, that is proportional to $\pi^*$, i.e.\
$
\pi^*(\x) = \frac{\pi(\x)}{Z}
$
where the normalization constant $Z$ is not necessarily known. 

We focus on particle-based methods. 
A \emph{particle-based approximation} of a discrete target distribution $\pi^*$ (with support $\X$) 
is simply another distribution $P$ with support $\X^P \subseteq \X$.
We sometimes refer to the probability of a particle as its weight. 

\textbf{Definition. }
We say, a discrete distribution $P$ (with support $\X^P \subseteq \X$)  
is the \emph{Optimal Particle-based Approximation of the Discrete Distribution $\pi^*$ on $\X^P$}  
(or simply the OPAD of $\pi^*$ on $\X^P$) if and only if, 
\begin{equation}
    \label{eq.optimal}
        P(\x) = \frac{\pi(\x)}{\sum\limits_{\V{x}' \in \X^P} \pi(\x')}, 
        \qquad \forall \x \in \X^P.
    \end{equation}

By Theorem~\ref{theo.main}, 
we will show that 
among all distributions, $P$, that can be defined on $\X^P$, 
the KL-divergence of $P$ from the target is minimum if and only if $P$ is the OPAD of the target (on $\X^P$).
The proof of Theorem~\ref{theo.main} relies on the following proposition.

\begin{proposition}[A variant of Jensen's inequality]
\label{theo.Jensen}
    For any discrete distribution, $P$, on a set, $\X^P$, 
    any integrable function $g$ and any 
    strictly convex function, $f: \mathbb{R} \to \mathbb{R}$,
    if $\exists \;\x_1, \x_2 \in \X^{P}$
such that $\x_1 \neq \x_2$ and $g(\x_1) \neq g(\x_2)$, then,
\begin{equation}
\label{eq.jensen1}
    \sum_{\x \in \X^{P}} f(g (\x)) P(\x) >
    f \left(\sum_{\x \in \X^{P}} g(\x) P(\x) \right).
\end{equation}
Otherwise, that is, if there exists a constant $c$
such that $\forall \x \in \X^P, g(\x) = c$, then, 
\begin{equation}
\label{eq.jensen2}
    \sum_{\x \in \X^{P}} f(g (\x)) P(\x) =
    f \left(\sum_{\x \in \X^{P}} g(\x) P(\x) \right) = f(c).
\end{equation} 

\end{proposition}
Proposition \ref{theo.Jensen} is a more nuanced variant of the (generalized) Jensen's inequality.  It deals with strictly convex functions, $f$, 
and refines its statement by distinguishing between the condition under which Jensen's inequality is strict and the condition that leads to equality.  
The proof of this proposition is provided in Appendix~\ref{sec.jensen}.

Our main theorem is as follows:

\begin{theorem}[Main Theorem]
\label{theo.main}
    Let $\pi^*$ be a discrete target distribution defined on support $\X$ and   
    associated with a score function $\pi$.
    Also, let $P$ be an arbitrary distribution with support $\X^P \subseteq \X$.
\begin{enumerate}
    \item The greatest lower bound on the Kullback–Leibler divergence of $P$ from the target, that is, $\KL{P}{\pi^*}$, is the negative log of the probability mass that the target assigns to the support of $P$: 
    \begin{equation}
    \label{eq.claim1}
        \text{infimum}\left( \KL{P}{\pi^*} \right) = -\log(\pi^*(\X^P)).
    \end{equation}
    \item This minimum divergence is reached if and only if:
    \begin{equation}
    \label{eq.claim2}
        P(\x) = \frac{\pi(\x)}{\sum\limits_{\V{x}' \in \X^P} \pi(\x')}, 
        \qquad \forall \x \in \X^P.
    \end{equation}
\end{enumerate}
\end{theorem}
{\it Proof. }
By letting $f(\cdot) := -\log(\cdot)$ (which is strictly convex) and 
$g(\cdot) := \frac{\pi^*(\cdot)}{P(\cdot)}$ (which is $P$-measureable for $P(\cdot) > 0$) 
in Proposition~\ref{theo.Jensen}, 
\begin{align*}
    \KL{P}{\pi^*} &:= \sum_{\x \in \X^{P}} -\log\left(\frac{\pi^*(\x)}{P(\x)}\right)P(\x) \\
    &\geq 
    -\log 
    \sum_{\x \in \X^{P}} \frac{\pi^*(\x)}{P(\x)}P(\x), \quad \text{ by \eqref{eq.jensen1}, \eqref{eq.jensen2}}
     \\
    &=
    -\log 
    \sum_{\x \in \X^{P}} \pi^*(\x)
    = -\log(\pi^*(\X^P)).
\end{align*}
Hence for all $P$ with support $\X^P \subseteq \X$, 
\begin{align}
 \KL{P}{\pi^*} \geq -\log(\pi^*(\X^P)),
\end{align}
which proves \eqref{eq.claim1}.

According to \eqref{eq.jensen2}
the equality, 
\[\KL{P}{\pi^*} = -\log(\pi^*(\X^P)),\]
happens
if and only if
$g(\x) := \frac{\pi^*(\x)}{P(\x)}=c$, for some constant $c > 0$.
Otherwise stated, the infimum on the KL-divergence is reached if and
and only if,
\begin{align*}
    P(\x) &\propto \pi^*(\x)   \\
    &= \frac{\pi^*(\x)}{\sum\limits_{\x' \in \X^P} \pi^*(\x')}, \text{ since $P$ is a distribution} \\
    &=  \frac{Z \cdot \pi^*(\x)}{\sum\limits_{\x' \in \X^P} Z \cdot \pi^*(\x')}
    = \frac{\pi(\x)}{\sum\limits_{\x' \in \X^P} \pi(\x')},
\end{align*}
which proves \eqref{eq.claim2}. 
\QED

\section{MCMC and OPAD}
\label{sect.opad+}

In this section, we focus on MCMC methods, because they are the most popular family of techniques to approximate discrete distributions.
In MCMC, the target $\pi^*$ (with support $\X$ and score function $\pi$) is approximated by a Markov chain $\tuple{\x_a^{[t]}}_{t=1}^{N}$ of $N$ states that belong to $\X$ but are not necessarily distinct.
The process is as follows: 
Starting from an initial state, $\x_a^{[1]} \in \X$, 
at each iteration, $t \in \{2, \ldots, N\}$, with Metropolis-Hastings acceptance probability, 
a proposal, $\x_p^{[t]}$, that is drawn from a proposal distribution, $q(\V{X} | \x_a^{[t-1]})$, is returned as the next state in the chain, $\x_a^{[t]} \eq \x_p^{[t]}$, 
otherwise, the previous state is repeated, $\x_a^{[t]} \eq \x_a^{[t-1]}$. \footnote{Note that the subscripts ``a" and ``p"  denote ``accepted" and ``proposed" states respectively.} 
The effective weight of each (distinct) state in the chain is proportional to the frequency of its occurrences in the chain. 
That is, the Markov chain 
$\tuple{\x_a^{[t]}}_{t=1}^{N}$
is equivalent to a distribution, $\PMC$: 
\begin{align*}
    \PMC(\x) := 
    \frac{
        \left\| \left\{ 
            t \text{ such that } 
            \x_a^{[t]} = \x
        \right\} \right\|
        }
    {N}, \qquad \forall \x \in \X^{\text{MC}}.
\end{align*}
where its support $\X^{\text{MC}}$, is the set of distinct states in the chain:
\[
\X^{\text{MC}} := \left\{ \x \text{ such that } \x \in \tuple{\x_a^{[t]}}_{t=1}^{N} \right\}.
\]
By theorem~\ref{theo.main}, 
\begin{align}
\label{eq:mc.less.opad}
   \KL{\PMC}{\pi^*} \geq \KL{P^{\text{OPAD}}_{\X^{\text{MC}}}}{\pi^*}, 
\end{align}
where $P^{\text{OPAD}}_{\X^{\text{MC}}}$ is the OPAD of $\pi^*$ on $\X^{\text{MC}}$.
Otherwise stated, 
the KL divergence of the approximation from the target is minimized
if the distinct states that are generated by the MCMC mechanism are 
weighted proportional to their target probability rather than the
frequency of their occurrence in the Markov chain.

In Section~\ref{sect:opad+} we show that the approximation error can be further reduced if we 
expand the set on which the OPAD approximation is constructed. 

\subsection{OPAD+}
\label{sect:opad+}

Let $\X^\text{MC+}$ be a set containing the initial state of the MCMC chain and the states that are proposed during the sampling process, regardless of whether they are accepted or rejected.  
\[
\X^\text{MC+} := 
\left\{\x_a^{[1]} \right\} \cup 
\left\{\x_p^{[t]} \text{ such that } t \in \{2, \ldots, N\} \right\}.
\]
Assuming that all proposals are drawn from the support of the target
and given that only a subset of the proposed samples are accepted,
\begin{align}
\X^\text{MC} \subseteq \X^\text{MC+} \subseteq \X.    
\end{align}
As such, by Corollary~\ref{cor.mass} and Inequality~\eqref{eq:mc.less.opad}, 
\begin{align}
\label{eq:geq.geq}
   \KL{\PMC}{\pi^*} \geq \KL{P^{\text{OPAD}}_{\X^{\text{MC}}}}{\pi^*}
    \geq \KL{P^{\text{OPAD}}_{\X^{\text{MC+}}}}{\pi^*}
\end{align}
in which $P^{\text{OPAD}}_{\X^{\text{MC+}}}$ is the OPAD of $\pi^*$ on $\X^\text{MC+}$ --
a distribution that from now on, we refer to as OPAD+.

Note that in \eqref{eq:geq.geq}, the equality only happens in an 
improbable case where (a) any proposed state is added to the Markov chain and (b) the frequency of occurrences of any state in the chain is exactly proportional to its target probability. 
In any other case, in terms of KL divergence from the target, OPAD+ outperforms the reference MCMC.

\begin{corollary}
\label{cor.mass}
Let $P_1$ and $P_2$ be OPADs of the target, $\pi^*$, on sets $\X^{P_1} \subseteq \X$ and $\X^{P_2} \subseteq \X$, respectively. 
If $\X^{P_1} \subset \X^{P_2}$
then 
$\KL{P_1}{\pi^*} > \KL{P_2}{\pi^*}$. 
\end{corollary}
{\it Proof. } Let $\X' := \X^{P_2} \backslash \X^{P_1}$.
\begin{align*} \footnotesize
    \KL{P_2}{\pi^*} &= -\log(\pi^*(\X^{P_2})), \qquad \text{ by \eqref{eq.claim1}} \\
    &= -\log(\pi^*(\X^{P_1} \sqcup \X')), \text{ by definition} 
    \\
    &= -\log\left(\pi^*(\X^{P_1}) + \pi^*(\X')\right) &&\text{} 
    \\
    &< -\log\left(\pi^*(\X^{P_1}) \right), \quad \text{since $\pi^*(\X') > 0$} 
    \\
    &= \KL{P_1}{\pi^*}. \hspace{33mm}\QED
\end{align*}

\begin{algorithm}[tb]
\caption{MCMC{\color{blue}/OPAD+}}
\textbf{Input:} $\x^{[1]}$, initial state $\in \X$;\\
\blank  \hspace{9mm} $\pi(\V{X})$, unnormalized target (on $\X$);\\
\blank  \hspace{9mm} $q(\V{X} | \x)$, proposal distribution (on $\X$);\\
\blank  \hspace{9mm} $N$, number of iterations;\\
\blank  \hspace{9mm} {\it\textsc{MC}}, set to $\top$ for MCMC {\color{blue} and to $\bot$ for OPAD+}.\\
\\
{\color{blue}$\X^{P} \eq \{\x^{[1]}\}$} \hspace*{\fill}\cmnt{Set of (distinct) proposals} \\
\textbf{For} {$t = 1, \ldots, N$} \textbf{do}\\
\blank  \hspace{4mm} $\x \sim q(\V{X} \,|\, \x^{[t]})$
\hspace*{\fill}\cmnt{Draw a proposal}\\
\blank  \hspace{4mm} {\color{blue}$\X^{P} \eq \X^{P} \cup \{\x\}$}\\
\blank  \hspace{4mm} \textbf{if}
$u \sim \text{Unif}(0, 1) \,<\,
\frac
    {\pi(\x) \cdot q\left(\x^{[t]} \,|\, \x\right)
    }
    {\pi(\x^{[t]}) \cdot q\left(\x \,|\, \x^{[t]}\right)
    } 
$
\textbf{ then }
$\x^{[t+1]} \eq \x$\\
\blank  \hspace{4mm} \textbf{else} 
$\x^{[t+1]} \eq \x^{[t]}$\\
\textbf{if} \textsc{MC} $\equiv \top$
{\bf then return} $\left\langle \x^{[t]} \right\rangle_{t=1}^N$
\hspace*{\fill}\cmnt{MCMC}\\
\textbf{else}\\
\blank  \hspace{4mm}
\color{blue}
$z = \sum\limits_{\x \in \X^{P}} \pi(\x)$ 
\hspace*{\fill}\cmnt{Normalization constant}\\
\blank  \hspace{4mm}
{\bf return} 
$\left\{ 
   \left\langle 
        \x, 
        \frac{\pi(\x)}{z}
    \right\rangle
    \text{ such that }
    \x \in \X^P 
\right\}$
\hspace*{\fill}\cmnt{OPAD+}\\
\label{alg:mcmc}
\end{algorithm}

Algorithm~\ref{alg:mcmc} illustrates an MCMC algorithm and its corresponding OPAD+.
In OPAD+,  the initial state and all the subsequent proposals are saved. 
At the end of the sampling process, these states are returned along with their weights which are proportional to their target probabilities.
Note that evaluating the (unnormalized) target probabilities is typically costly. 
However, in practice, for each proposed node, this evaluation only needs to be done once when the acceptance probability is computed and then saved. This means that OPAD+ does not introduce any significant computational overhead.

\section{Experiments}
\label{sect.exper}

Our theoretical results in the previous section ensure that, the KL divergence of OPAD and OPAD+ from the target,
is less than any particle-based approximation that uses the same particles. 
In this section, we measure this improvement in 
famous applications of discrete distributions, 
including Ising models \citep{propp1996exact},
Bayesian Variable Selection 
\citep{Mitchell1988,Edward1993,ni2023handbook}
and Bayesian Structure Learning \citep{Kuipers2017,suter2021bayesian}.
We limit the scope of our experiments to 
relatively low-dimensional models where 
the exact computation of the KL divergence is feasible.

\begin{figure*}[th]
     \centering
     \begin{subfigure}[b]{0.45\textwidth}
         \centering
         \includegraphics[width=\textwidth]{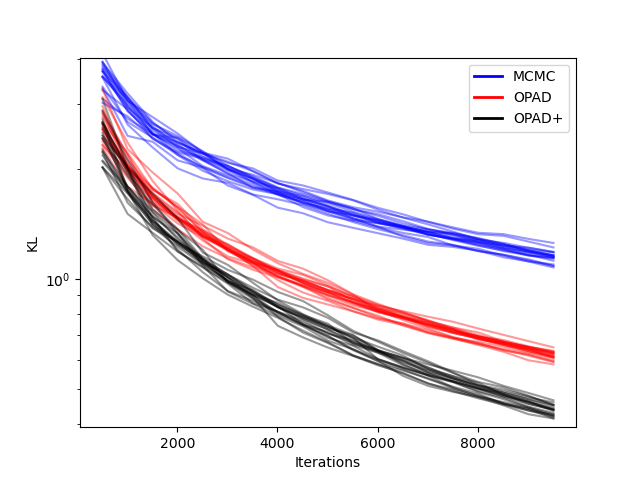}
         \caption{}
         \label{fig:ising.trace.short}    
     \end{subfigure}
     \begin{subfigure}[b]{0.45\textwidth}
         \centering
         \includegraphics[width=\textwidth]{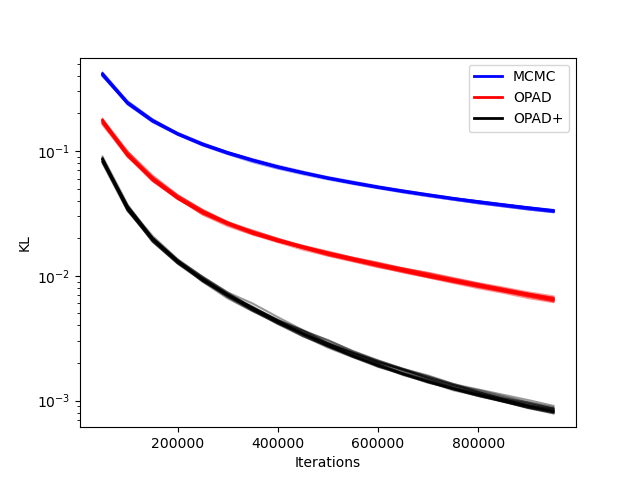}
         \caption{}
         \label{fig:ising.trace.long}
     \end{subfigure}
     \caption{KL divergence from the 1D Ising model (Equation~\ref{eq.ising1d}) versus sampling iterations: Plotted for 20 chains of the reference MCMC (see Section~\ref{sect.ising1d.mcmc}) (blue curves), as well as their OPAD and OPAD+ counterparts (red and green curves)
     (a) for 10K sampling iterations and (b) for 1 million iterations. 
     }
     \label{fig:Ising}
\end{figure*}

\subsection{1D Ising Model}
\label{sect.ising}
Our first experimental target distribution, $\pi(\x)$, 
is a 1D periodic Ising model on a closed loop of size $m$.
In this model, $\X=\{-1, 1\}^m$
and the probability of each state $\x := (x_1, \ldots, x_m) \in \X$, is given by a Boltzmann distribution,
$\pi(\x) \propto \exp^{-\beta H(\x)}$,
with inverse temperature parameter $\beta \geq 0$ and the Hamiltonian function, $H(\x)$,
\begin{equation}
\label{eq.ising1d}
    H(\x) := -\sum_{j=1}^m J_j x_j x_{j+1} - \mu \sum_{j=1}^m h_j x_j,
\end{equation}
where $x_{m+1} = x_1$. 
The magnetic moment, $\mu$,  
and the interaction strengths, $J_j$, 
and 
internal magnetic field strengths, $h_j$,
are model parameters that we set as follows:
$\mu=1$,
$\forall j\in \{1, \ldots m\}$, $J_j=1$ and $h_j = 0.1$. 
We also let $m=15$ and $\beta=0.5$.

\subsubsection{Reference MCMC for Ising Model}
\label{sect.ising1d.mcmc}
The MCMC proposal generation mechanism that is suitable for this Ising model is as follows:
a site (index) $j^*$ is chosen uniformly,
$j^* \sim \text{Unif}(\{1, \ldots, m\})$,
and its spin is flipped. That is,
if the current state is 
$\x^c := (x_1^c, \ldots, x_m^c)$,
then the proposal will be 
$\x^p := 
(x_1^p, \ldots, x_m^p)$,
in which,
\begin{equation*}
    x_j^p = 
    \begin{cases}
    -x_j^c& \text{, if } j=j^*\\
    x_j^c& \text{, otherwise}
\end{cases},
\quad \forall j \in \{1, \ldots m\}.
\end{equation*}

\subsection{Bayesian Variable Selection}
\label{sect.bvs}
In linear regression models, 
the relation between scalar responses, $y_i \in \R$, and predictor vectors,
$\x_i := (x_{i,1}, \ldots, x_{i,m})^\top \in \R^m$,
is assumed to be linear plus some independent Gaussian noise.
For $n$ points $\{(\x_i, y_i)\}_{i=1}^n$:
\begin{align}
    \label{eq.lin.reg}
    \V{y} = \V{X} \V{\beta} + \V{\epsilon}
\end{align}
where $\y:=(y_1, \ldots, y_n)^\top \in \R^n$ is the vector of responses;
$\V{X} := (\x_1, \ldots \x_n)^\top \in \R^{n \times m}$ is the design matrix; 
$\V{\beta} = (\beta_1,...,\beta_m)^\top \in \R^m$
is the vector of model coefficients 
and $\V{\epsilon} \sim \mathcal{N}(\V{0}, \sigma^2 \V{I}_n)$
is the vector of i.i.d. noise values.  We assume, without loss of generality, that the responses and predictors have zero means.
In Bayesian Variable Selection, 
the goal is to identify important predictors of the response. One method of achieving this is to allow a model co-efficient to be identically zero, by placing a mixture distribution on the prior of the coefficients, where a component of that mixture is a point mass at zero, \citep{Mitchell1988, Edward1993, SMITH1996317}. Such a prior is often referred to as a {\it Spike and Slab} prior. 

To do this, a binary vector is introduced, 
called the {\it selection indicator vector}, 
$\V{\gamma} := (\gamma_1, \ldots, \gamma_m)^\top \in \{0, 1\}^m$, where for $1\leq j \leq m$, $\gamma_j = 1$ if and only if 
$\beta_j \neq 0$. We define the sets $S_k := \{j \text{ such that } \gamma_j = k\}$ for $k=0,1$  to
be the set of indices of the included predictors ($k=1$), and excluded predictors ($k=0$). An example of such a prior is 
\begin{eqnarray*}
    \V{\beta}_{S_1} | S_1, \sigma^2, \V{X}_{S_1} &\sim &
    \mathcal{N}\left(\V{0}, g \sigma^2 
    \left(\V{X}_{S_1}^\top \V{X}_{S_1} \right)^{-1}\right),\\
    \V{\beta}_{S_0} &\sim &\delta(\V{0}),
\end{eqnarray*}
with $g>>1$ and where $\delta(\V{0})$ is the Dirac delta function evaluated at zero. If in addition, we place an Inverse-gamma prior on $\sigma^2$ with 
parameters $a=3$ and $b=1$:
\begin{align*}
    \sigma^2 &\sim \text{IG}(a, b),
\end{align*} 
then the marginal likelihood, $p(\y | \V{\gamma}, \V{X})$  becomes 
\begin{align}
\notag
    &p(\y | \V{\gamma}, \V{X}) 
    \propto 
    \frac{1}{ 
(g+1)^{\frac{\|S_1\|}{2}}
} 
\\
&\cdot
\left[
    \frac{
        \y^\top \left( \V{I} - 
        \frac{g}{g+1}
        \V{X}_{S_1}\left(\V{X}_{S_1}^\top \V{X}_{S_1}\right)^{-1} \V{X}_{S_1}^\top \right) \y + 2b }
        {2}
\right]^{-a -\frac{n}{2}},
\label{eq.bvs.like}
\end{align}
where $\|S_1\|$ is the cardinality of $S_1$, (see  for example \cite{Mitchell1988,Edward1993,SMITH1996317}).\\
To complete the model specification we assume, apriori,  that elements of $\V\gamma$ are i.i.d Bernoulli$(\rho)$ so that
\begin{align}
\label{eq.prior.gamma}
    p(\V{\gamma}) \propto 
    \rho^{\|S_1\|} \cdot (1-\rho)^{\|S_0\|}.
\end{align}
In this setting, the target distribution is 
the posterior distribution of $\V{\gamma}$, 
\begin{align}
    \label{eq:bvs.target}
    p(\V{\gamma} | \y, \V{X}) 
    &\propto  
    p(\y | \V{\gamma}, \V{X}) p(\V{\gamma}),
\end{align}
which is  the product of terms \eqref{eq.prior.gamma} and \eqref{eq.bvs.like}.

\begin{figure*}[th]
     \centering
     \begin{subfigure}[b]{0.45\textwidth}
         \centering
         \includegraphics[width=\textwidth]{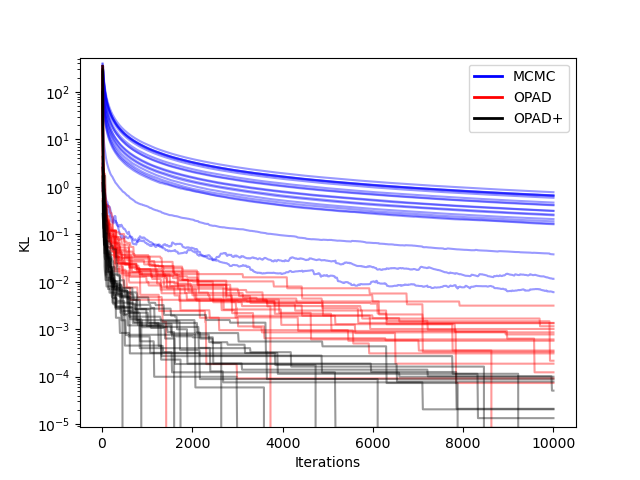}
         \caption{}
         \label{fig:syn.bvs.trace}    
     \end{subfigure}
     \begin{subfigure}[b]{0.45\textwidth}
         \centering
         \includegraphics[width=\textwidth]{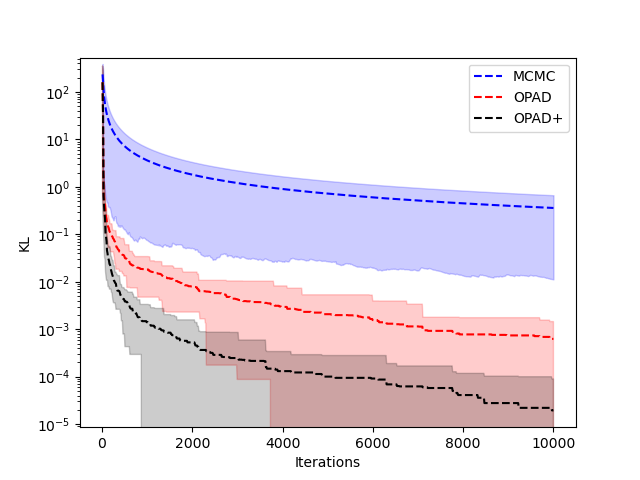}
         \caption{}
         \label{fig:syn.bvs.conf}
     \end{subfigure}
     \\
     \begin{subfigure}[b]{0.45\textwidth}
         \centering
         \includegraphics[width=\textwidth]{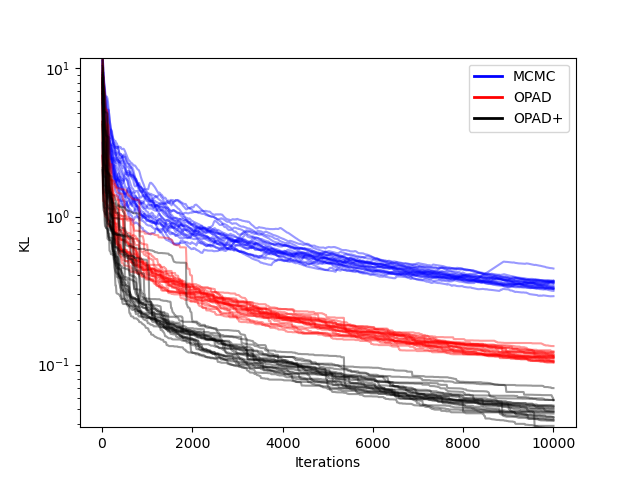}
         \caption{}
         \label{fig:mice.bvs.trace}    
     \end{subfigure}
     \begin{subfigure}[b]{0.45\textwidth}
         \centering
         \includegraphics[width=\textwidth]{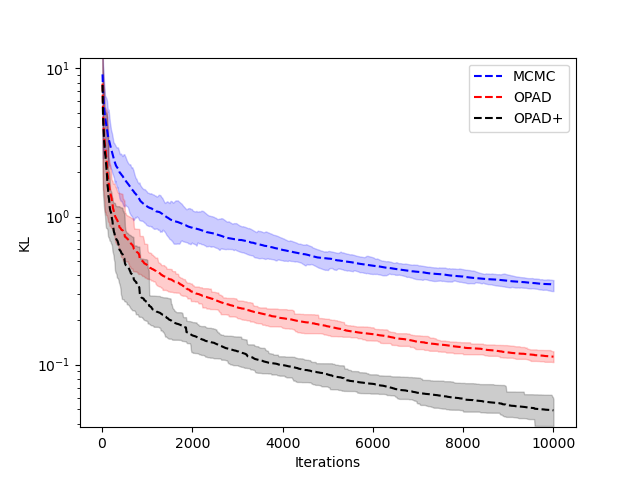}
         \caption{}
         \label{fig:mice.bvs.conf}
     \end{subfigure}
     \caption{KL divergence from the Bayesian Variable Selection target distribution (Equation~\ref{eq:bvs.target}) 
     versus sampling iterations using a reference MCMC explained in 
     Section~\ref{sect.bvs.mcmc}:
     On the left, the divergence of 20 MCMC chains is plotted (blue curves), 
     along with their corresponding OPAD and OPAD+ counterparts (red and black curves), 
     whereas on the right, the mean and 95\% confidence interval of these algorithms are plotted. 
     In (a) \& (b), the target posterior is w.r.t.\ 200 synthetic data points
     (see Section~\ref{sect.bvs.synthetic.data}) while in 
     (c) \& (d), the target posterior is w.r.t.\ the real-world Mice nutrition dataset (Section~\ref{sect.bvs.real.data}).
     }
     \label{fig:bvs}
\end{figure*}

\subsubsection{Reference MCMC for Bayesian Variable Selection}
\label{sect.bvs.mcmc}
The MCMC proposal is constructed by flipping a randomly chosen element of the current state. That is,
if the current state is 
$\V{\gamma}^c := (\gamma_1^c, \ldots, \gamma_m^c)^\top$,
then the proposal will be 
$\V{\gamma}^p := 
(\gamma_1^p, \ldots, \gamma_m^p)^\top$,
in which,
\begin{equation*}
    \gamma_j^p = 
    \begin{cases}
    1-\gamma_j^c& \text{, if } j=j^*\\
    \gamma_j^c& \text{, otherwise}
\end{cases},
\quad \forall j \in \{1, \ldots m\},
\end{equation*}
with
$j^* \sim \text{Unif}(\{1, \ldots, m\})$.

\subsubsection{Synthetic dataset}
\label{sect.bvs.synthetic.data}
We construct a $20$-dimensional synthetic ground true model and then draw observational data from it as follows:

1. We let $m=20$ and construct a ground true selection indicator vector 
$\V{\gamma}^\star:=(\gamma_1^\star, \gamma_2^\star, \ldots, \gamma_m^\star)^\top$ 
by drawing $\gamma_{j}^\star$ 
from a Bernoulli distribution with parameter $\rho = 0.5$.

2. We construct a ground true model coefficient vector 
$\V{\beta}^\star := (\gamma_1^\star \alpha_1, \ldots, \gamma_m^\star \alpha_m)^\top$ where $\alpha_j \sim \text{Unif}(-4, 4)$.

3. To generate $n=200$ data points, 
we construct an $n \times m$ design matrix 
$\V{X}$ where 
its elements are uniformly drawn from the interval $[-3, 3]$ and rescaled so that the mean is equal to 0.

4. Finally, the observed response vector, $\y$, is generated by equation \eqref{eq.lin.reg} in which we let the noise variance $\sigma^2=1$.

\subsubsection{Real-world dataset}
\label{sect.bvs.real.data}
We use `Mice Nutrition' data set described in \cite{SOLONBIET2014418}, 
where the response variable, $y$,
is `age at death (w)' and 
the (original) predictor variables includes:\\
1. `Dry weight food eaten (g/mouse/cage/d)' ($D$),\\
2. `Cellulose intake (g/d)' ($Ce$),\\
3. `P eaten (kJ/mse/cage/d)' ($P$),\\
4. `C eaten (kJ/mse/cage/d)' ($C$),\\
5. `F eaten (kJ/mse/cage/d)' ($F$), and\\
6. `Energy intake (kJ/mse/cage/d)' ($E$).\footnote{
$P$, $C$, and $F$ stand for mice protein, carbon, and fat intakes (per cage per day), respectively. 
} 
\\
To allow the model to capture non-linear relationships, we add the following 
\emph{interaction terms} to the predictors:\\
7. $P \times C$,
8. $P \times F$,
9. $C \times F$,
10. $P \times C \times F$,
11. $\frac{P}{P + C + F}$,
12. $\frac{C}{P + C + F}$, and
13. $\frac{F}{P + C + F}$.\\
Each predictor is standardized to have zero mean and unit variance. 

\subsection{Bayesian Structure Learning}
\label{sect.bsl}
Bayesian Structure Learning is the task of  
creating (and if necessary, approximating) a posterior distribution over all \emph{Bayesian Networks} (BNs), that can be defined on a set of random variables, $\V{X}:= (X_1, \ldots, X_n)$, 
conditioned on observational data \citep{friedman2003being}. 

More specifically, let $G$ be a \emph{Directed Acyclic Graph} (DAG) 
that establishes parent/child relationships between the nodes $X_1$ to $X_n$.
It corresponds to a 
\emph{Bayesian Network} (BN) that represents a factorization of the joint probability $p(X_1, \dots, X_n)$ into the following product of conditional distributions:
\[
p(X_1, \dots, X_n | G) = \prod_{i=1}^{n} p(X_i | \text{Pa}_G(X_i))
\]
where $\text{Pa}_G(X_i)$ denotes the parents of node $X_i$ according to DAG, $G$.

The observational data, $\cD := \{\V{x}^{(d)}\}_{d=1}^N$, (with $\V{x}^{(d)} := (x^{(d)}_1, \ldots x^{(d)}_n)$) includes $N$ realisations of $\V{X}$; and, the likelihood function is:
\begin{align}\label{eq:dag}
    p(\cD | G) &= \prod_{d=1}^N p(\V{X} = \V{x}^{(d)} | G)  \\
    &= \prod_{d=1}^N \prod_{i=1}^n p \left( X_i = x^{(d)}_i | \text{Pa}_G(X_i) = \text{Pa}_G(x^{(d)}_i) \right), \nonumber
\end{align}
where $\text{Pa}_G(x^{(d)}_i)$ denotes the corresponding realisation of $\text{Pa}_G(X_i)$ in $\V{x}^{(d)}$.
The likelihood functions that are often used in the literature
includes 
(a) the \emph{Bayesian Dirichlet equivalent score} (BDe) 
\citep{heckerman2013learning}
with a Dirichlet parameter prior for binary and categorical data, and (b) the
\emph{Bayesian Gaussian equivalent score} (BGe) 
\citep{geiger2002parameter} 
with an inverse Wishart prior for continuous
data.
In this setting, the \emph{target distribution} is the posterior distribution of DAGs,
\begin{align}
\label{eq:bsl.posterior}
p(G | \cD) \propto p(G) p(\cD | G).    
\end{align}


\subsubsection{Reference MCMC for Bayesian Structure Learning}
\label{sect.bsl.mcmc}
In this setting, we use two reference MCMC algorithms:
 Structure MCMC 
\citep{madigan1995structure} and 
Partition MCMC \citep{Kuipers2017}\footnote{Python code for structure and partition MCMC is based on \url{https://github.com/annlia/partitionMCMC}. }.
 In structure MCMC, the proposal generation mechanism involves adding, deleting, or reversing a single edge from the current DAG in the MCMC chain.
 Partition MCMC is more sophisticated and internally conducts sampling in the space of \emph{node partitions} that represent partial orderings of nodes.  
The proposed moves in the partition space consist of joining adjacent subsets or splitting an existing subset into two. At each sampling iteration, given the current partition, a DAG that is compatible with that partial ordering is sampled and returned. 

\subsubsection{Synthetic dataset}
To generate observational data, we follow the approach of \cite{Lorch2021} and \cite{yu2019dag}. 
We Let $n=5$ and per MCMC chain, randomly draw a ground true Erd\H{o}s--R\'enyi DAG \citep{erdos1960evolution} with an expected vertex degree $d \in \{1,2,3\}$. 
For each ground true DAG, we draw the parameters of the valid edges from a $U[0,2]$. Subsequently, we simulate 200 data points from the linear DAG model defined in \eqref{eq:dag}.
Based on this synthetic observational data, in our experiments, we construct a unique DAG posterior distribution per MCMC chain, using Equation~\eqref{eq:bsl.posterior}.

\begin{figure*}[t]
     \centering

     \begin{subfigure}[b]{0.32\textwidth}
         \centering
         \includegraphics[width=\textwidth]{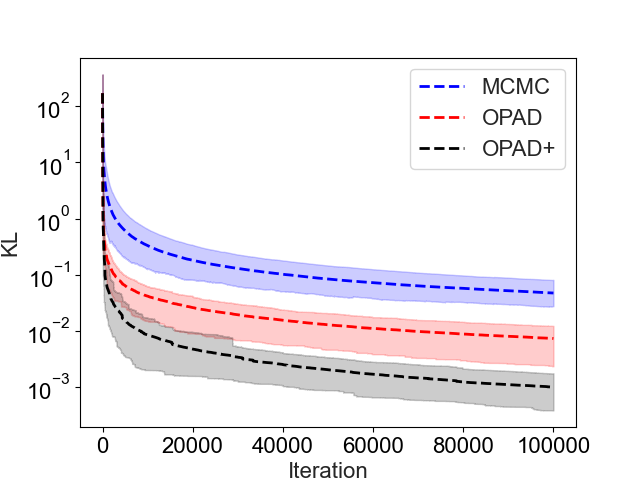}
         \caption{$\text{ER}(5, 1)$}
         \label{fig:strucutre_learning_er_1}    
     \end{subfigure}
     \begin{subfigure}[b]{0.32\textwidth}
         \centering
         \includegraphics[width=\textwidth]{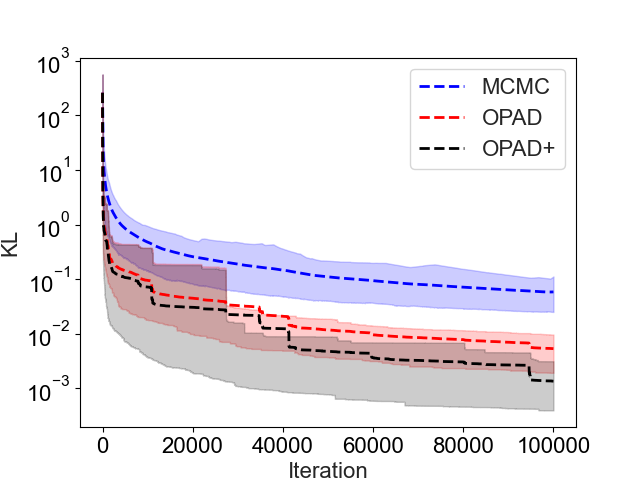}
         \caption{$\text{ER}(5, 2)$}
         \label{fig:strucutre_learning_er_2}
     \end{subfigure}
     \begin{subfigure}[b]{0.32\textwidth}
         \centering
         \includegraphics[width=\textwidth]{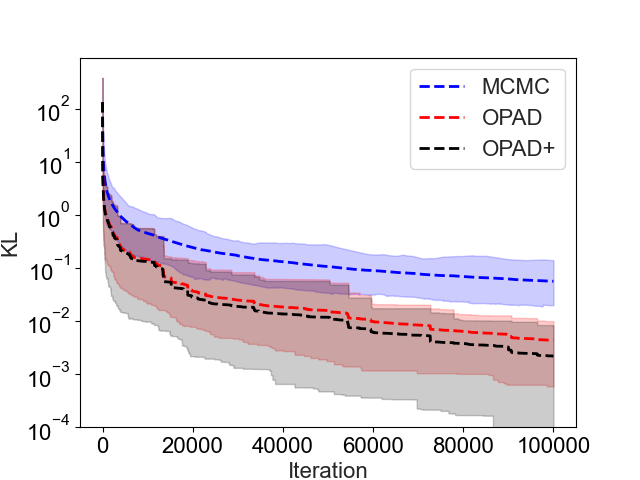}
         \caption{$\text{ER}(5,3)$}
         \label{fig:strucutre_learning_er_3}
    \end{subfigure}
    \caption{KL divergence from the Bayesian Structure Learning target distribution (Equation~\ref{eq:bsl.posterior}) versus sampling iterations where the reference MCMC is Structure MCMC. The target posterior is synthesized per MCMC
    chain using 200 data points generated from a randomly constructed Erd\H{o}s--R\'enyi ground truth DAG, denoted as $\text{ER}(n, d)$, where $n$ represents the number of nodes and $d$ indicates the expected vertex degree. The mean and 95\% confidence intervals of 20 chains are plotted.}
     \label{fig:strucutreMCMC}
\end{figure*}

\begin{figure*}[t]
     \centering

     \begin{subfigure}[b]{0.32\textwidth}
         \centering
         \includegraphics[width=\textwidth]{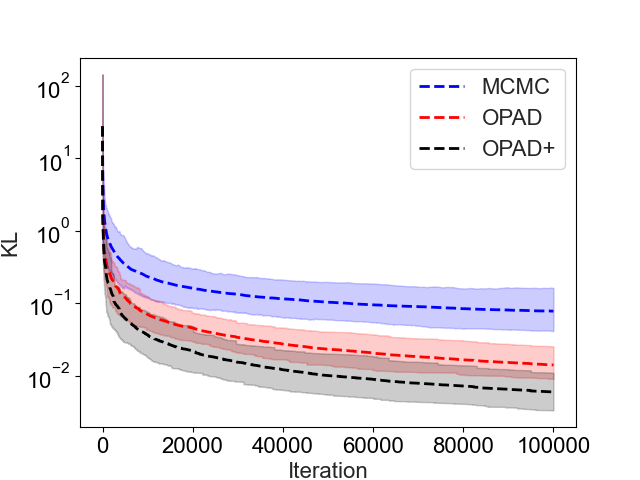}
         \caption{$\text{ER}(5, 1)$}
         \label{fig:strucutre_learning_er_1}    
     \end{subfigure}
     \begin{subfigure}[b]{0.32\textwidth}
         \centering
         \includegraphics[width=\textwidth]{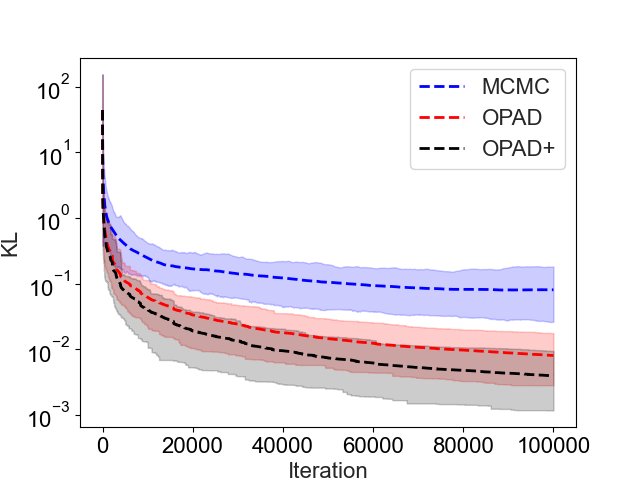}
         \caption{$\text{ER}(5, 2)$}
         \label{fig:strucutre_learning_er_2}
     \end{subfigure}
     \begin{subfigure}[b]{0.32\textwidth}
         \centering
         \includegraphics[width=\textwidth]{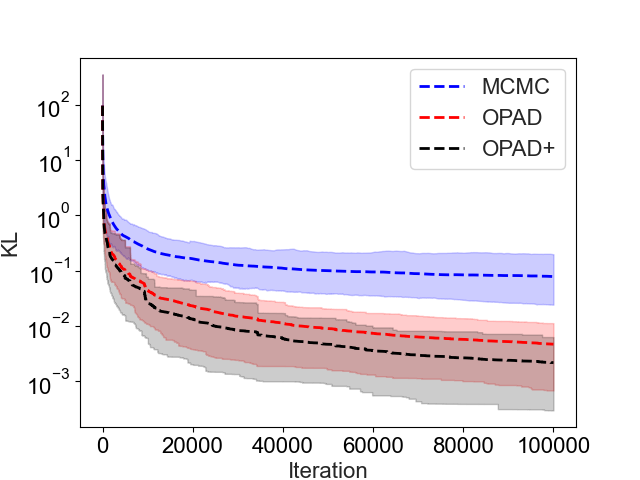}
         \caption{$\text{ER}(5,3)$}
         \label{fig:strucutre_learning_er_3}
    \end{subfigure}
    \caption{
    KL divergence from the Bayesian Structure Learning target distribution (Equation~\ref{eq:bsl.posterior}) versus sampling iterations where the reference MCMC is Partition  MCMC. The target posterior is synthesized per MCMC
    chain using 200 data points generated from a randomly constructed Erd\H{o}s--R\'enyi ground truth DAG, denoted as $\text{ER}(n, d)$, where $n$ represents the number of nodes and $d$ indicates the expected vertex degree. The mean and 95\% confidence intervals of 20 chains are plotted.}
     \label{fig:partitionMCMC}
\end{figure*}

\subsection{Evaluations}

For each experimental model, 
we evaluate the performance of the relevant reference MCMC algorithm(s) and their  corresponding OPAD/OPAD+ algorithms 
by plotting their KL divergence from the target distribution versus the number of drawn samples (iterations).

In each setting, these plots are depicted for 20 independent MCMC chains.
The initial point of each chain is drawn uniformly from the space of possible states. 

In Figures~\ref{fig:ising.trace.short} 
and \ref{fig:ising.trace.long}
the target distribution is the Ising model introduced in Section~\ref{sect.ising} and the sampling is carried on for 10K and 1M iterations respectively. 
The results show that OPAD and OPAD+ significantly outperform the reference and the difference is more noticeable in higher iterations.
The remaining experimental plots are depicted for 10K iterations only. 

Figure~\ref{fig:bvs}, 
deals with the Bayesian Variable Selection models 
of Section~\ref{sect.bvs}.
The target,  
the posterior distribution of the \emph{selection indicator vector}, 
is constructed with respect to synthetic (top) and real-world (bottom) observational datasets.
Along with the divergence measures associated with individual chains (left), the mean and 95\% confidence interval of the chains are also plotted for each algorithm (right).

Figures~\ref{fig:strucutreMCMC} and \ref{fig:partitionMCMC} 
deal with the Bayesian Structure Learning setting of Section~\ref{sect.bsl}.
In Figure~\ref{fig:strucutreMCMC}, the 
reference MCMC is Structure MCMC 
whereas in Figure~\ref{fig:partitionMCMC}, 
the reference is Partion MCMC (see Section~\ref{sect.bvs.mcmc}).
As before, the mean and 95\% confidence intervals are plotted for MCMC, OPAD, and OPAD+.

These experiments indicate that in different models, the proposed OPAD and OPAD+ methods consistently and significantly outperform the reference MCMC algorithms in the sense that their KL divergence from the target is much lower (from one-tenth to one-thousandth of the corresponding MCMC). 
Additionally, our experiments suggest that as sampling continues for more iterations, the relative performance of OPAD and OPAD+ with respect to the associated MCMC remains stable or improves.
Our results also show that OPAD+ consistently outperforms OPAD
and in most experiments, the difference between these two variants of the algorithm is significant. This highlights the importance of the information that can be gathered from the rejected MCMC proposals to improve the approximation.  



\section{Conclusion}
In this paper, we prove that there is a unique particle weighting mechanism that minimizes the KL divergence of any particle-based method from a discrete target distribution, leading to an \emph{Optimal Particle-based Approximation of the Discrete Target Distribution} (OPAD). These optimal weights are simply proportional to the target probabilities of the particles, normalized to sum to one. As such, through a simple reweighting that does not incur any costly computations, the approximation error of any particle-based method, such as various Markov Chain Monte Carlo (MCMC) algorithms, can be reduced.

If an existing particle-based method computes the target scores of states that are not included in the chosen particles (e.g., the proposed MCMC states that are rejected), those states can also be incorporated into the corresponding OPAD to create the OPAD+ variant of the algorithm. This leads to a further decrease in the KL divergence of the resulting approximation from the target distribution.

The quantitative improvement of OPAD and OPAD+ over their respective particle-based methods is evaluated on well-known families of discrete distributions, including the Ising model, Bayesian Variable Selection, and Bayesian Structure Learning. Our experiments demonstrate that through OPAD/OPAD+ particle reweighting, the divergence from the target can be reduced by several orders of magnitude.

It is important to emphasize that our proposed OPAD/OPAD+ method is not a substitution for MCMC or other particle-based methods that approximate discrete distributions. Rather, it complements and improves them by optimally incorporating the information that they compute.  
The primary advantage of existing particle-based techniques lies in their ability to identify probability modes and retrieve regions of space where probability mass is concentrated. This is particularly crucial in high-dimensional models where evaluating all states is infeasible. 
Our findings show that simply by dropping the repeated states 
and reweighting them according to the OPAD mechanism, 
the optimal particle-based approximation (that can be made with these states) is created.

\bibliographystyle{apalike}
\bibliography{refs}

\newpage

\newpage
\appendix 
\onecolumn

\section{Proof of proposition 1}
\label{sec.jensen}
\textbf{
\!\!Proposition 1 [A variant of Jensen's inequality].}
\textit{
    For any distribution, $P$, on a set, $\X^P$, 
    any $P$-measureable function $g$ and any 
    strictly convex function, $f: \mathbb{R} \to \mathbb{R}$,  
} 
\begin{equation}
\label{eq.jensen_}
    \sum_{\x \in \X^{P}} f(g (\x)) P(\x) \geq 
    f \left(\sum_{\x \in \X^{P}} g(\x) P(\x) \right),
\end{equation}
and the equality occurs if and only if
$
    \forall \x \in \X^P, g(\x) = c, 
$
where $c$ is an arbitrary constant. 

{\it Proof. }
Let $T_n$ abbreviate the fact or assumption that the theorem holds for all
sets, $\X^P$, with $n$ elements.  
In a proof by induction, we first establish $T_2$ (base case).
Subsequently, we show that if $T_n$ holds then $T_{n+1}$ also holds.

\vspace{1mm}
\textsc{Base Case} ($T_2$).
Consider the case where the cardinality of $\X^P$ is 2. 
Let $\X^P := \{\x_1, \x_2\}$. 
Since $f$ is strictly convex, 
by definition, 
$\forall y_1, y_2 \in \mathbb{R}$ and $\forall \rho \in [0,1]$:
\begin{align}
\rho f(y_1) + (1 - \rho) f(y_2) &> f(\rho y_1 + (1-\rho) y_2), 
&&\text{ if and only if }
y_1 \neq y_2, 
\label{eq.convexity.gr}
\\
\rho f(y_1) + (1 - \rho) f(y_2) &= f(\rho y_1 + (1-\rho) y_2), 
&&\text{ if and only if }
y_1 = y_2. 
\label{eq.convexity.eq}
\end{align}

By letting $\rho := P(\x_1)$ 
(which entails, $(1-\rho) = P(\x_2)$), 
$y_1 := g(\x_1)$ and $y_2 := g(\x_2)$, 
equations \eqref{eq.convexity.gr} and \eqref{eq.convexity.eq}
become: 
\begin{align*}
 \sum_{\x \in \{\x_1, \x_2\}} f(g (\x)) P(\x) &> 
    f \left(\sum_{\x \in \{\x_1, \x_2\}} g(\x) P(\x) \right), 
\text{ if and only if }
g(\x_1) \neq g(\x_2), \text{ by \eqref{eq.convexity.gr}} 
\\
\sum_{\x \in \{\x_1, \x_2\}} f(g (\x)) P(\x) &=
    f \left(\sum_{\x \in \{\x_1, \x_2\}} g(\x) P(\x) \right), 
\text{ if and only if }
g(\x_1) = g(\x_2), \text{ by \eqref{eq.convexity.eq}} 
\end{align*}
which establishes $T_2$.
\vspace{1mm}
\\
\textsc{Induction Step}.\\
Assuming $T_n$, i.e. that the theorem holds for
sets, $\X^P$, with $n$ elements, 
we prove $T_{n+1}$ i.e. we show that the theorem 
also holds for sets $\X^P$, with $n+1$ elements:
\\
Let $\X^P := \{\x_1, \ldots, \x_n, \x_{n+1}\}$,
$\rho_i$ denote $P(\x_i)$ and $y_i$ denote $g(\x_i)$.
\begin{align}
    \sum_{\x \in \X^{P}} f(g (\x)) P(\x)
    &= 
    \rho_1 f(y_1)  + \ldots +  \rho_{n-1} f(y_{n-1})
    +
    \rho_{n} f(y_{n}) + \rho_{n+1} f(y_{n+1})
    \notag
    \\
    &=
    \rho_1 f(y_1)  + \ldots +  \rho_{n-1} f(y_{n-1})
    +
    (\rho_{n} + \rho_{n+1})
    \left(
    \frac{\rho_{n}}{\rho_{n} + \rho_{n+1}} f(y_{n}) + 
    \frac{\rho_{n+1}}{\rho_{n} + \rho_{n+1}} f(y_{n+1})
    \right)
    \notag
    \\
    &\geq
    \rho_1 f(y_1)  + \ldots +  \rho_{n-1} f(y_{n-1})
    +
    (\rho_{n} + \rho_{n+1})
    f\left(
    \frac{\rho_{n}}{\rho_{n} + \rho_{n+1}} y_{n} + 
    \frac{\rho_{n+1}}{\rho_{n} + \rho_{n+1}} y_{n+1}
    \right)
    \label{eq.assume.theorem2}
    \\
    &\geq
    f 
    \left(
    \rho_1 y_1  + \ldots +  \rho_{n-1} y_{n-1}
    +
    (\rho_{n} + \rho_{n+1})
    \left(
    \frac{\rho_{n}}{\rho_{n} + \rho_{n+1}} y_{n} + 
    \frac{\rho_{n+1}}{\rho_{n} + \rho_{n+1}} y_{n+1}
    \right)
    \right) 
    \label{eq.assume.theorem_n}
    \\
    \notag
    &=
    f 
    \left(
    \rho_1 y_1  + \ldots +  \rho_{n-1} y_{n-1}
    +
    \rho_{n} y_{n} + 
    \rho_{n+1} y_{n+1}  
    \right) = 
f \left(\sum_{\x \in \X^{P}} g(\x) P(\x) \right),   
\end{align}
in which inequalities 
\eqref{eq.assume.theorem2} and \eqref{eq.assume.theorem_n}
hold by $T_2$ and $T_n$ assumptions respectively.
Therefore \eqref{eq.jensen_} holds for $n+1$ elements. 
The above relations also show that the equality, 
\begin{align*}
    \sum_{\x \in \X^{P}} f(g (\x)) P(\x) = 
    f \left(\sum_{\x \in \X^{P}} g(\x) P(\x) \right),
\end{align*}
holds 
if and only if both \eqref{eq.assume.theorem2} and \eqref{eq.assume.theorem_n} are equalities
which according to $T_2$, and $T_n$,  is the case if and only if, $y_1 = \ldots = y_{n+1}$. That is, 
$$g(\x_1) = \ldots = g(\x_n) = g(\x_{n+1}),$$
which establishes $T_{n+1}$.
\QED

\twocolumn

\end{document}